\documentclass[10pt,a4paper]{article}

\usepackage{amsmath}
\usepackage{amssymb}
\usepackage{graphicx}
\usepackage{url}
\usepackage{comment}
\usepackage[numbers,sort&compress]{natbib}
\usepackage{enumitem}
\usepackage{setspace}
\usepackage[left=1in,right=1in,top=1in,bottom=1in]{geometry}
\usepackage{subfigure}

\title{Theoretical Analysis of Power-law Transformation on Images for Text Polarity Detection}

\author{\bf Narendra Singh Yadav and Pavan Kumar Perepu\\Indian Institute of Information Technology, Sri City, India\\Email: \{narendrasingh.y@iiits.in, pavankumar.p@iiits.in\}}
\date{}

\begin{document}

\maketitle

\begin{abstract}
Several computer vision applications like vehicle license plate recognition, captcha recognition, printed or handwriting character recognition from images etc., text polarity detection and binarization are the important preprocessing tasks. To analyze any image, it has to be converted to a simple binary image. This binarization process requires the knowledge of polarity of text in the images. Text polarity is defined as the contrast of text with respect to background. That means, text is darker than the background (dark text on bright background) or vice-versa. The binarization process uses this polarity information to convert the original colour or gray scale image into a binary image. In the literature, there is an intuitive approach based on power-law transformation on the original images. In this approach, the authors have illustrated an interesting phenomenon from the histogram statistics of the transformed images. Considering text and background as two classes, they have observed that maximum between-class variance between two classes is increasing (decreasing) for dark (bright) text on bright (dark) background. The corresponding empirical results have been presented. In this paper, we present a theoretical analysis of the above phenomenon.  
\end{abstract}

{\bf keywords:} Image binarization, Text polarity, Image histogram, Power-law transformation, Between-class variance

\section{Introduction}
\label{intro}
In most of the computer vision and pattern recognition problems \cite{pami2014, elagouni2012, textdetectcnn2014, milyaev2015ijdar, DKE2018, ivcdeep2020} like printed and handwriting character recognition, number plate recognition etc., binarization is a primary preprocessing task. Binarization or thresholding is a process of converting a colour or grayscale image into a binary image. To perform binarization, there is an imperative need to determine text polarity. In any printed or handwritten document, text can be darker on bright background or vice-versa. This polarity information is required for image binarization. For example, consider two images shown in Fig. \ref{plcurve}. First image has bright text ({\bf 7996}) on dark background while the second one has dark text ({\bf 7966}) on bright background. The corresponding binary images are shown in Fig. \ref{polarityissue}. In the first image, the text pixels are white and black in the second one. If computer vision or pattern recognition algorithms need to process text pixels, this polarity information is required. 

Binarization algorithms \cite{nc2017, nc2019, nikolaos2015, ieeesys2013, Chaki2014} are classified mainly into two types: Global and Local. Global algorithms are based on the histograms of the complete image while local ones are based on the histograms of local regions in the image. Otsu's method \cite{otsu} is a traditional and standard global binarization algorithm. It converts the original image into a binary image based on a threshold such that all pixels with values less than (greater than) a threshold are given black (white) colour. This process generates black text pixels on white background. If pixels with values less than (greater than) a threshold are given white (black) colour, white text on black background is generated. This Otsu's approach has still been applied and extended in recent works \cite{shi2016otsu}. 

As mentioned above, text polarity has to be detected before the binarization process. Several approaches \cite{shi2016otsu, jayant2007, icip2008} have been proposed in the literature to determine the text polarity. Most of these approaches are based on hard thresholds and are not robust. We have also previously proposed a deep learning approach to determine text polarity \cite{ppk2021}. However, there is a simple and intuitive approach based on Otsu's approach itself to detect text polarity \cite{shi2016otsu} which will be discussed in the next section. In this method, authors have observed an interesting phenomenon after applying the power-law transformation on the original images. This phenomenon is useful in identifying the polarity, and it has been empirically presented and illustrated. 

In this paper, we present a preliminary theoretical analysis of the power-law transformation-based text ploarity detection. The paper is organized as follows: Section \ref{textpolarity} presents Otsu's based text polarity detection method. In Section \ref{analysis}, we present the theoretical analysis of the above approach and the paper is concluded in Section \ref{conc}.


\section{Text Polarity Detection Method}
\label{textpolarity}

Otsu's binarization method \cite{otsu, shi2016otsu} computes a threshold from an image histogram such that text pixels are isolated from the background pixels. If an image has $L$ gray levels, this method divides the histogram (gray level value versus frequency/probability) into two classes, text and background, based on a threshold value that isolates the two classes. This threshold is chosen in such a manner that between-class variance is maximized. This approach is briefly discussed below \cite{shi2016otsu}.

Let $p_i$ be the probability of gray level, $i$, which is computed by the ratio of number of pixels with gray level, $i$, to the total number of pixels. Let $\omega_1$ and $\omega_2$ denote the two classes (text and background, or vice-versa) such that $\omega_1$ and $\omega_2$ denote the gray values in the range, $[1,t]$ and $[t+1,L]$, respectively. 

Probability of classes, $\omega_1$ and $\omega_2$, at threshold, $t$, denoted by, $w_1(t)$ and $w_2(t)$, are given by:

\begin{center}
$w_1(t)= \displaystyle \sum_{i=1}^t p_i$ and $w_2(t)= \displaystyle \sum_{i=t+1}^L p_i$. 
\end{center}

Mean of classes, $\omega_1$ and $\omega_2$, at threshold, $t$, denoted by, $\mu_1(t)$ and $\mu_2(t)$, are given by:

\begin{center}
$\mu_1(t)= \displaystyle \frac{\sum_{i=1}^t i p_i}{w_1}$ and $\mu_2(t) = \displaystyle \frac{\sum_{i=t+1}^L i p_i}{w_2}$.
\end{center}

Total mean of the entire image, $\mu_T = \displaystyle \sum_{i=1}^{L}ip_i$.

Between-class variance at threshold, $t$, denoted by $\sigma_B^2(t)$, is given by:

\begin{center}
$\sigma_B^2(t) = w_1(t) (\mu_1(t)-\mu_T)^2 + w_2(t) (\mu_2(t)-\mu_T)^2=w_1(t)w_2(t)(\mu_1(t)-\mu_2(t))^2.$
\end{center}

Otsu's method \cite{otsu} chooses an optimal threshold, $t^*$, that maximizes the between-class variance, as given below. 

\begin{center}
$t^* = \displaystyle \underset{t}{\text{argmax }} \sigma_B^2(t) $
\end{center}

\begin{figure}[t]
\centering
\includegraphics[scale=0.6]{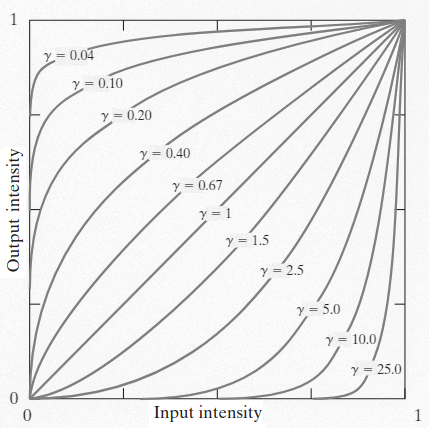}
\caption{Power-law transformation, $o=i^\gamma$, where $i$ and $o$ are input and output intensity values respectively, for different values of $\gamma$ \cite{gonzalez} }
\label{powerlaw}
\end{figure}

Let $\sigma_B^2(t^*)$ be the corresponding maximum between-class variance ($MBCV$). Let a power-law transformation \cite{gonzalez, shi2016otsu} be applied on the original image. If $i$ is a gray-level intensity value in the original image, then its intensity value in the transformed output image, is given by, $o = i^\gamma$, where $\gamma$ is a parameter. Power-law transformation functions \cite{gonzalez} for different values of $\gamma$, are shown in Fig. \ref{powerlaw}. For the analysis, input intensity values in the range, $[0, L]$, are generally normalized \cite{shi2016otsu} to the range, $[0, 1]$. It can be observed from this figure, if $\gamma < 1$, initial narrow range of input intensity values (dark region in the input) are mapped to a broad range of output intensity values. This generates an output image with  high contrast. Similarly, a broad range of input pixel values are mapped to a narrow range of output intensity values (dark region in the output), if $\gamma >1$. In this case, an output image with low contrast is generated. 

\begin{figure}[t]
\includegraphics[scale=0.75]{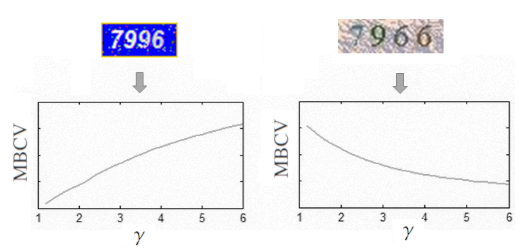}
\caption{MBCV vs Power ($1/\gamma$) curve for two types of text polarity: Bright text on dark background (left) and Dark text on bright background (right).}
\label{plcurve}
\end{figure}

In \cite{shi2016otsu}, authors have observed an interesting phenomenon on the value of $MBCV$ of the power-law transformed image. $MBCV$ values for different values of $\gamma$ can be computed and plotted. In their work, they have normalized the input intensity values in the range, $[0, 1]$, and applied the following transformation: $o = i^{1/\gamma},\,i \in [0,1]$. As shown in Fig. \ref{plcurve}, $MBCV$ values increase (decrease) as the value of $\gamma$ increases (that means, $1/\gamma$ values decrease, shown as upper curves above $\gamma =1$, in Fig. \ref{powerlaw}), for bright (dark) text on dark (bright) background. This $MBCV$ vs $\gamma$ plot can be used to detect text polarity. 

\begin{figure}
\begin{minipage}[t]{0.45\textwidth}
\centering
\includegraphics[scale=0.7]{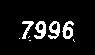}
\end{minipage}
\hspace{0.3cm}
\begin{minipage}[t]{0.5\textwidth}
\centering
\frame{\includegraphics[scale=0.8]{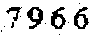}}
\end{minipage}
\caption{Binarized images for the original images shown in Fig. \ref{plcurve}. In the left image, white pixels (black pixels) form text (background) and vice-versa in the right image. }
\label{polarityissue}
\end{figure}

As shown in Fig. \ref{polarityissue}, Otsu's method converts the original image into a binary image by converting all pixels below (above) the optimal threshold to black (white). As power-law transformation gives the text polarity information, any computer vision application can process the text pixels accordingly, which are white (black) in the left (right) image of Fig. \ref{polarityissue}.

In \cite{shi2016otsu}, authors have only empirically illustrated this phenomenon based on power-law transformation. In our current work, we have presented a detailed theoretical analysis which will be discussed in the next section. 
\section{Theoretical Analysis}
\label{analysis}
Consider the following expression for the maximum between-class variance ($MBCV$) for the optimal threshold, $t^*$, on the original image:

\begin{equation}
\sigma_B^2(t^*) = w_1(t^*)w_2(t^*)(\mu_1(t^*)-\mu_2(t^*))^2
\label{mbcvexp}
\end{equation}

If the power-law (with power, $((1/\gamma)<1)$) is applied on the original image, darker intensity values are stretched to a broader range of output values, which increases the image contrast (discussed above).  In \cite{shi2016otsu}, authors have empirically observed that $MBCV$ values decrease (increase) as $\gamma$ value increases (decreases) for dark (bright) text on bright (dark) background. In this work, we have proposed that this phenomenon is possible under some conditions based on the nature of the image histogram and contrast, for both the cases discussed below. We have also shown and discussed some counter examples which do not satisfy these conditions.

\subsection{Case I: Bright text on dark background}
Let the original image has bright text on dark ground. Let the power-law transformation with power less than one is applied on the original image, and its $MBCV$ value is computed. It can be clearly observed from equation \ref{mbcvexp}, this $MBCV$ value is less than that of the original image if the original image satisfies the following properties:
\begin{enumerate}[label=(\alph*)]
\item $w_1(t^*) \approx w_2(t^*)$ and
\item $t^*$ is closer to 0, and $\mid \mu_1(t^*) - \mu_2(t^*)\mid < \epsilon$, where $\epsilon$ is a small value.
\end{enumerate}

First condition ensures that the image has a bi-modal histogram with two objects (text and background) in almost equal proportions. That means, between-class variance depends on the difference between class means, $\mu_1$ and $\mu_2$, as $w_1$ and $w_2$ are constants (almost equal). Second condition ensures that objects are closer to each other (low contrast) towards darker side. If the original image satisfies these properties, as per the power-law transformation, darker pixels are stretched, which also increases the distance between the class means, $\mu_1(t^*)$ and $\mu_2(t^*)$. According to equation \ref{mbcvexp}, this increases $MBCV$ (as shown in Fig. \ref{plcurve}, left figure).

\begin{figure}
\centering
\subfigure[]{\includegraphics[width=0.4\textwidth]{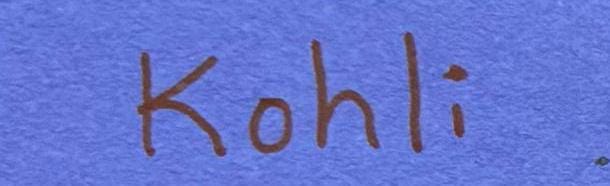}}
\subfigure[]{\includegraphics[width=0.3\textwidth]{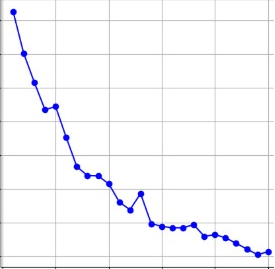}}
\subfigure[]{\includegraphics[width=0.3\textwidth]{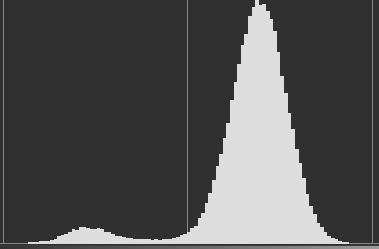}}
\caption{(a) Original Image (bright text on dark background) (b) $MBCV$ vs $\gamma$ curve (c) Histogram}
\label{bod1}
\end{figure}

Consider a bright on dark text image shown in Fig. \ref{bod1} (a), which does not satisfy the above properties. Its $MBCV$ vs Power ($1/\gamma$) curve and histogram are shown in Figs. \ref{bod1} (b) and (c), respectively. It can be observed that the histogram is almost uni-modal and so the $MBCV$ vs Power plot has fluctuating (and mostly decreasing) trend.

\subsection{Case II: Dark text on bright background}
Let the original image has dark text on bright ground. Let the power-law transformation with power less than one is applied on the original image, and its $MBCV$ value is computed. It can be clearly observed from equation \ref{mbcvexp}, this $MBCV$ value is greater than that of the original image if the original image satisfies the following properties:
\begin{enumerate}[label=(\alph*)]
\item $w_1(t^*) \approx w_2(t^*)$ and
\item $t^*$ is closer to 1, and $\mid \mu_1(t^*) - \mu_2(t^*)\mid > \Delta$, where $\Delta$ is a large value.
\end{enumerate}

First condition ensures that the image has a bi-modal histogram with two objects (text and background) in almost equal proportions. Second condition ensures that objects are away from each other (high contrast) towards brighter side. If the original image satisfies these properties, as per the power-law transformation, darker pixels are stretched while brighter pixels almost remain constant. That means, the class mean, $\mu_1(t^*)$, moves towards $\mu_2(t^*)$. According to equation \ref{mbcvexp}, this decreases $MBCV$ (as shown in Fig. \ref{plcurve}, right figure).

Consider a dark text on bright background image shown in Fig. \ref{bod1} (a), which does not satisfy the above properties. Its $MBCV$ vs Power ($1/\gamma$) curve and histogram are shown in Figs. \ref{bod1} (b) and (c), respectively. It can be observed that the histogram is almost uni-modal and so the $MBCV$ vs Power plot has fluctuating (and mostly increasing) trend.

\begin{figure}
\centering
\subfigure[]{\includegraphics[width=0.4\textwidth]{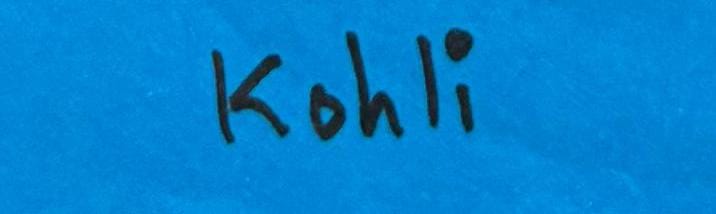}}
\subfigure[]{\includegraphics[width=0.3\textwidth]{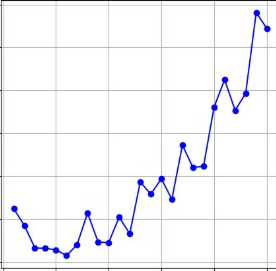}}
\subfigure[]{\includegraphics[width=0.3\textwidth]{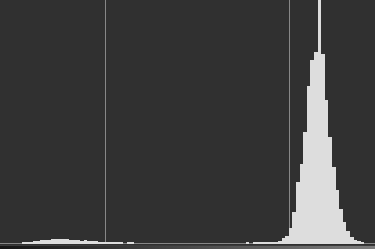}}
\caption{(a) Original Image (dark text on bright background) (b) $MBCV$ vs $\gamma$ curve (c) Histogram}
\label{bod2}
\end{figure}

\section{Conclusions}
\label{conc}
In this paper, we focus on text polarity detection in various document images. This approach is based on applying the power-law transformation on the original images. In the literature, it has been empirically observed that the $MBCV$ values of the transformed image increase (decrease) as the values of $\gamma$, increase, for bright (dark) text on dark (bright) background. A preliminary theoretical analysis on the above phenomenon has been presented. We have discussed the conditions under which this phenomenon is possible along with counter examples. In future, we will extend this analysis to the images with both the polarities.

\section{Acknowledgements*}
Figures 4(a) and (b), and 5(a) and (b) are generated by the project students, Mr. Ritvik and Mr. Krishna Vardhan.

\bibliographystyle{apalike} 
\bibliography{Powerlaw_TextPolarity}

\end{document}